# Multi-Stage Transmission Line Flow Control Using Centralized and Decentralized Reinforcement Learning Agents

Xiumin Shang[1], Jingping Yang[2], Bingquan Zhu[3], Lin Ye[3],
Jing Zhang[3], Jianping Xu[2], Qin Lyu[3], Ruisheng Diao[1,*]

1 AI & System Analytics, GEIRI North America, San Jose, CA 95134
2 System Operation, Jinhua Electric Power Company, Jinhua, Zhejiang, China
3 System Operation, SGCC Zhejiang Electric Power Company, Hangzhou, Zhejiang, China
* `ruisheng.diao@geirina.net` *

## Abstract

Planning future operational scenarios of bulk power systems that meet security and economic constraints typically requires intensive labor efforts in performing massive simulations. To automate this process and relieve engineers' burden, a novel multi-stage control approach is presented in this paper to train centralized and decentralized reinforcement learning agents that can automatically adjust grid controllers for regulating transmission line flows at normal condition and under contingencies. The power grid flow control problem is formulated as Markov Decision Process (MDP). At stage one, centralized soft actor-critic (SAC) agent is trained to control generator active power outputs in a wide area to control transmission line flows against specified security limits. If line overloading issues remain unresolved, stage two is used to train decentralized SAC agent via load throw-over at local substations. The effectiveness of the proposed approach is verified on a series of actual planning cases used for operating the power grid of SGCC Zhejiang Electric Power Company.

## 1 Introduction

Secure and economic operation of modern power grids is a complex control problem, which requires various constraints being met at all time including voltage profiles, frequencies, line flows, etc. To better plan for future operational scenarios, massive simulation studies using grid models are typically conducted to analyze the risks at both normal conditions and under contingencies (loss of one or more grid element, a.k.a., N-1/N-k). Once security issues are identified, control measures can be applied to mitigate such issues. However, given the complexity, nonlinearity and high dimensionality of bulk power grids, deriving feasible and optimal controls to meeting security and reliability standards is quite challenging, which typically requires many hours of manual tuning and adjustment of controllers while running simulation studies. With the increased penetration of intermittent energy, more uncertain and stochastic behaviors are being observed in modern power systems, causing even more challenges for power engineers in finding effective and low-cost control measures. Thus, a automated process to achieve the above goals is much needed by the power community.

---

*Dr. Ruisheng Diao is with GEIRI North America as Deputy Department Head, AI & System Analytics, leading the development of a number of AI-based applications in power systems (autonomous voltage control, line flow control, intelligent maintenance scheduling for power utilities and control centers). Cell phone: (480)-414-7095, website: https://www.linkedin.com/in/ruisheng-diao-ph-d-pe-789a9655/

Recent research on power flow control was focused on limiting inter-area tie-line flows for market operation [1, 2, 3], or resolving overloading issues across the entire power network for security reasons [4, 5, 6]. The lack of effective methods to find feasible solutions under all uncertainties like load changes and contingencies makes power flow control quite challenging, especially for large-scale power network. Deep Reinforcement Learning (DRL), as a fast growing sequential decision making approach in recent years, has demonstrated effectiveness on solving large-scale complex system control problem [7, 8, 9]. Applying this approach to grid control problems has been a promising direction, and different applications have been reported, such as automating the emergency control process [10], regulating system voltage profiles and transmission line flows at normal and contingency conditions [11, 12, 13].

In this work, a multi-stage reinforcement learning-based method is presented to automatically tune grid controllers to mitigate transmission line flow violation problems at base and contingency conditions. This control problem is first formulated as MDP where observation space contains line flows, bus voltages, generator outputs and loads of a power network; and control space is formed by selected generators via active power adjustment and substation loads via shifting among local substation groups. Inside each load control group, the sum of load values remain constant (active power consumption). A two-stage RL framework is developed to train centralized agent for generator control; and decentralized agent for substation loads control. A software prototype is developed and tested on actual power grid planning cases used for grid operation by the SGCC Zhejiang Electric Power Company.

## 2 Problem Formulation

### 2.1 Background

#### 2.1.1 Power Grid Physics

To maintain the safety of power grid, quasi-steady-state conditions need to be met at all time. To represent such behavior, power components including generators, buses, loads, transmission lines, and transformers need to be modeled as algebraic equations, with their practical equality constraints and inequality constraints.

$$\begin{aligned}\sum P^g - \sum P^d - gV^2 = \sum P \\ \sum Q^g - \sum Q^d - bV^2 = \sum Q\end{aligned} \quad (1)$$

Eq 1 gives the equality constraints, it represents the active power $P$ and reactive power $Q$ balance of each bus (or node) in the power network, indicating the sum of injected power must be equal to the sum of outflow power at all time to maintain power system security. $V$ is voltage magnitude of a node, $g$ and $b$ are the ratio parameters.

$$\begin{aligned}P^{min} \leq P \leq P^{max}, Q^{min} \leq Q \leq Q^{max}, \\ \sqrt{P^2 + Q^2} \leq S^{max}, V^{min} \leq V \leq V^{max},\end{aligned} \quad (2)$$

The inequality constraints in Eq 2 stand for physical limits of various power components, requiring that all line flows, generator outputs and voltage magnitudes operate within their limits. $S$ is complex power.

#### 2.1.2 Reinforcement Learning

Reinforcement learning is usually modeled as Markov Decision Process (MDP) to describe how an agent learn the desired behaviors by sequentially interacting with its surrounding environment. The MDP includes global state space $S$, action space $A$, transition probability $P$, reward function $R$. At each time step $t$, the agent observes a state $s_t \in S$, takes an action $a_t \in A$, and receives a scalar reward $r(s_t, a_t)$. The agent's decision making behavior is defined by a policy $\pi : P(A) \leftarrow S$, which maps states to a probability distribution over actions. The performance of the agent is evaluated by Q value, the average future accumulated reward of taking the current action, $Q = E[\sum_{n=0}^{N} \gamma^n r_n]$. The agent's objective is to find a policy which could maximize this reward.

Among different RL approaches, we choose Soft Actor-Critic (SAC) [14] algorithm, which has demonstrated state-of-art performance on both sample efficiency and stability due to its unique capability of maximizing both expected rewards and entropy during the training process.

$$J_Q(\theta) = E_{(s_t,a_t) \sim D}[\frac{1}{2}Q_\theta(s_t, a_t) - (r(s_t, a_t) + \gamma V_\psi(s_{t+1}))^2] \tag{3}$$

$$J_\pi(\phi) = E_{s_t \sim D, \epsilon \sim N}[\alpha \log(\pi_\phi(a_t|s_t)) - Q_\theta(s_t, a_t)] \tag{4}$$

Eq 3 and 4 are the loss functions of action-value (Q) and policy network ($\pi$), $\theta$ and $\phi$ are the deep network parameters accordingly. $V$ is the state value function, and $\psi$ is its network parameter. Using stochastic gradient decent, the loss function try to minimize the difference between our estimated value $Q_{\theta(s_t,a_t)}, \alpha \log(\pi_\phi(a_t|s_t))$ and our network target $(r(s_t, a_t) + \gamma V_\psi(s_{t+1}))^2, Q_\theta(s_t, a_t)$.

## 2.2 State and Action space

This section describes the state and action space for both generator control and load control. For stage one, the generator control, state space is defined as a vector $S_g = (P, V, G)$, where $P$ represents a set of line flows within the study zone; $V$ is voltage magnitude of buses in the same zone; and $G$ stands for a vector of generator active power outputs; the action space $A_g$ is defined by $G$, as control signals to regulate power flow on transmission lines in the power grid. For stage two, the load control, state space is defined as a vector $S_d = (P, V, D)$, $D$ stands for a vector of load power; the action space $A_d$ is defined by $D$, as control signals for load redistribution. The batch normalization techniques is applied for both state and action values during training to maintain consistency of various units.

## 2.3 Design of Reward

This section describes the reward calculation process. We assume the power system has $N$ transmission lines, if the system could maintain secure and reliable when losing any one transmission line, we call this power system $N - 1$ secure, and the control problem is called $N - 1$ contingencies[15]. As our purpose is to derive feasible operating conditions that can survive various contingencies in the power grid against potential line flow overloading issues, the definition of reward function for generator control and load control are the same.

The reward function is defined as the sum of base reward $r_{base}$ and contingency reward $r_{con}$ in Eq 5.

$$r = r_{base} + r_{con} \tag{5}$$

The base reward $r_{base}$ is to check the security of the fully connected power grid, and Eq 6 shows the calculation process of accumulating overload power among $N$ transmission lines. The contingency reward $r_{con}$ is to measure how secure the power system is after cutting certain transmission lines from the grid network, Eq 7 describes that for each cutting line $k$, the rest of $N - 1$ lines will be calculated for how much they have exceeded their line limits.

$$r_{base} = a \sum_{k=1}^{N} \Big[ max(|P_{from}|, |P_{to}|) - bP_{limit} \Big] \tag{6}$$

$$r_{con} = a \sum_{k=1}^{N} \sum_{l=1}^{N-1} \Big[ max(|P_{from}|, |P_{to}|) - bP_{limit} \Big] \tag{7}$$

Here, $P_{from}$ and $P_{to}$ are active power measurements at the "from end" and "to end" of a transmission line. $P_{limit}$ is the physics attribute of this line, showing its thermal or stability limit. Considering the power grid physics constraints based on Eq 1 and Eq 2, the amount difference of $max(|P_{from}|, |P_{to}|)$ and $P_{limit}$ is accumulated to the reward function as a signal to regulate the power flow passing through transmission lines. $a$ is a reward scalar, $b$ is a impact ratio of the line capacity. We use $\frac{1}{3500}$ and 0.9 respectively in our experiments. $l$ is one of the rest of $N - 1$ lines, and $k$ is one of the total lines $N$.

## 3 Multi-Stage DRL Control Approach

### 3.1 System Design

The system design of our approach is shown in Figure 1, it includes "offline training" and "online using" process.

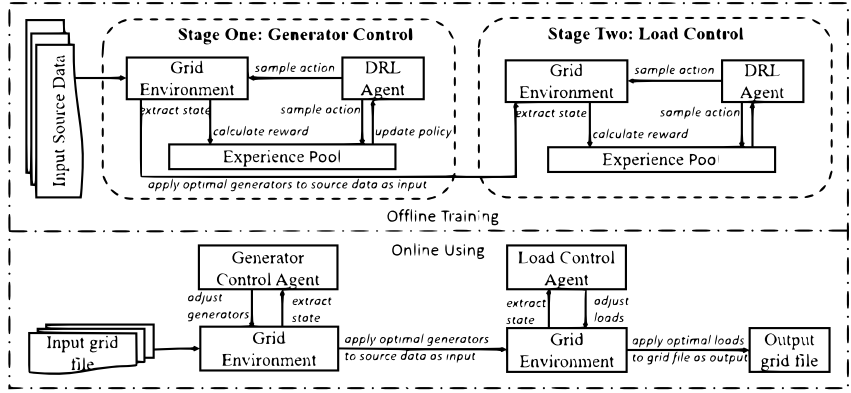

Figure 1: Multi-stage transmission line flow DRL control framework

For offline training process, there are two stages: generator control stage and load control stage. Both stages includes three components, the "Grid Environment", "DRL Agent" and "Experience Pool". The "Grid Environment" component takes the source power grid data as input to start the DRL training process, and it extracts state information from the data file. It also runs a power flow numerical solver to calculate reward function as described from Eq 5, to collect state $s_t$ and reward $r$ for the current time step. The "DRL Agent" component updates the agent's policy and sample action $a_t$ from the policy. The "Experience Pool" component collects $(s_t, a_t, r, s_{t+1})$ from the other two components as sample data for updating policy and Q value function networks. After finishing generator control stage, the power grid network will be re-evaluated by the power flow numerical solver. If line overloading issue still exists, the optimal generator values found by the optimal policy will be applied to power grid data and be used as input data for stage two load control.

For "online using" process, the input grid file will be processed by stage one generator control to adjust the generator active power values. After the first stage completes, the optimal generator values found by the agent will be applied to the grid file. For the second stage, the updated grid file will be used as input to the grid environment. After this stage finish, the optimal load values will be found and be applied to the grid file. This grid file is the final state of grid network, which could stay secured against contingency.

### 3.2 Algorithm Implementation

The implementation of our proposed approach is shown in Algorithm 1, it is called Multi-Stage DRL Power Flow Control algorithm (MSDRL). Line $1 \sim 13$ shows the training process of getting the control policy for generators under contingency. Line $15 \sim 28$ shows the training process for load control policy if the overloading issue could not be solved by only running generator control process. For generator control, line $7\sim10$ generates the Markov decision tuple $(s_t, a_t, r, s_{t+1})$ for updating policy and value function networks. Line $11\sim13$ shows that when the agent collects sample data tuples more than batch_size, the policy and Q functions network will perform a stochastic update based on Eq 3 and Eq 4. The load control process is similar to generator control process.

## 4 Experiments and Results

### 4.1 Environment

The experiment environment used in our study is power grid planning models, which are generally used for creating future operational scenarios of a real power system. There are more than 6,500 buses,

**Algorithm 1:** Multi-Stage DRL Power Flow Control Alogirthm (MSDRL)

    **Generator Control :**
1. initialize policy network $\phi$, $Q$ network $\theta_1, \theta_2$ with random weights ;
2. initialize replay buffer $D$ as empty dictionary ;
3. collect contingency lines $L_c$ cause failure;
4. **for** *each episode* **do**
5.     reset environment and collect initial state $s_t$;
6.     **for** *each step* **do**
7.         sample action $a_t$ from policy $\pi(a_t|s_t)$;
8.         DRL agent interacts with power grid environment;
9.         calculate reward $r$ with eq.xx under different training conditions;
10.        collect next state $s_{t+1}$ and send tuple $(s_t, a_t, r, s_{t+1})$ to $D$;
11.        **if** *step > batch_size* **then**
12.            sample a batch tuples $\{(s_i, a_i, r, s_{i+1})\}$ from $D$;
13.            update policy and $Q$ networks with Eq. XX ;
14. output optimal policy $\phi_{gen}$ for generator control ;
15. **if** *overload still exist* **then**
16.     detect overloading areas as action space;
    **Load Control :**
17.     initialize policy network $\phi$, $Q$ network $\theta_1, \theta_2$ with random weights ;
18.     initialize replay buffer $D$ as empty dictionary ;
19.     collect target loads as action space;
20.     **for** *each episode* **do**
21.         reset environment and collect initial state $s_t$;
22.         **for** *each step* **do**
23.            sample action $a_t$ from policy $\pi(a_t|s_t)$;
24.            DRL agent interacts with power grid environment;
25.            calculate reward $r$ with eq.xx under different training conditions;
26.            collect next state $s_{t+1}$ and send tuple $(s_t, a_t, r, s_{t+1})$ to $D$;
27.            **if** *step > batch_size* **then**
28.                sample a batch tuples $\{(s_i, a_i, r, s_{i+1})\}$ from $D$;
29.                update policy and $Q$ networks with Eq. XX ;
30.     output optimal policy $\phi_{load}$ for load control

600 generators, 6,000 lines and 4,300 transformers in the original planning models. Two experiments has been conducted in the interested zone of Jinhua, Zhejiang province of East China. For the first experiment, the zone of interest inside the planning model has 224 buses, 231 transmission lines and 7 generators, representing January of 2019 conditions; for the second experiment, the planning model used for testing contains 226 buses, 222 transmission lines and 7 generators, representing October of 2019 conditions.

### 4.2 Results and Analysis

In the first experiment, for stage one generator control, the state space dimension is 462 and the action space dimension is 7 based on the given planning model. The result of training iteration steps v.s. episodes is shown in Figure 2. The grid network has been successfully converged during generator control process, and there is no overloading issue remain.

In the second experiment, the state space dimension is 455 and the action space dimension is 7 for stage one generator control. After finishing the training process of stage one, there is no optimal policy reached which means adjusting generator values alone could not find feasible solutions. It is because of the control limitation of the selected 7 generators, that they could not cover the entire grid area in regulating line flows. Thus, a load control process needs to be applied to local unsolved area. In this model, there are 6 substation loads has been identified as control objective for stage two load control, the state space dimension is 453 and the action space dimension is 5 with the 6th load being used as a swing load absorbing changes from other 5 loads. After applying load control, the overloading issue has been resolved successfully. The result is shown in Figure 3.

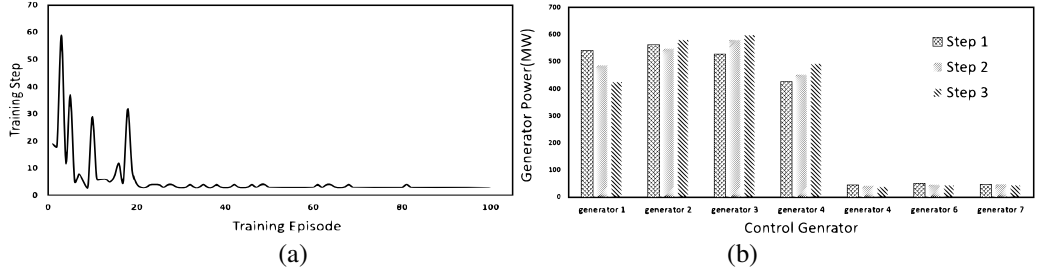

(a)  (b)

Figure 2: Results of generator control stage in first experiment as feasible solution could be found in stage one. (a) shows the offline training process, the training steps decrease along the training episodes increase direction, and it converges after around 20 episodes; (b) shows the online using process, the generators reach the optimal values in only three steps.

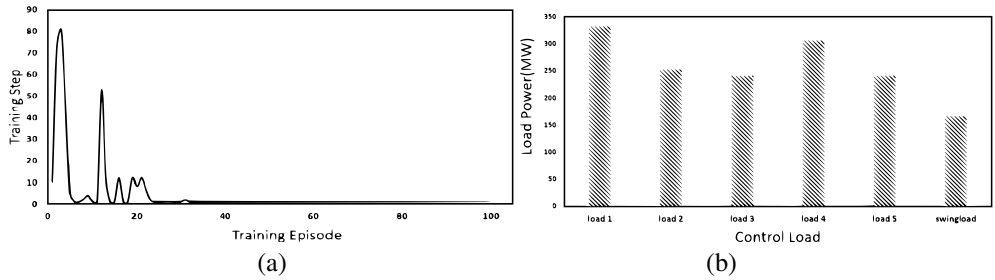

(a)  (b)

Figure 3: Results of load control stage in second experiment as feasible solution has been found in stage two. (a) shows the offline training process, the training steps decrease along the training episodes increase direction, and it converges after around 30 episodes; (b) shows the online using process, the loads reach the optimal values in only one step.

Both experiments show the effectiveness of solving transmission line flow control problem under contingencies with our multi-stage DRL transmission line flow control approach.

## 5  Conclusion

In this paper, a novel approach is proposed to solve power grid planning and operation problem considering various contingencies using a multi-stage DRL based framework. The proposed method is tested on two real power grid models, indicating this framework can help automatically search for feasible power grid operating solutions considering uncertainties. The current work focuses on investigating a small city-level area for load adjusting during second stage. This method could be further expanded to different local areas using the second stage control in future work.

**Broader Impact**

In this work, we present a multi-stage methodology of training centralized and decentralized reinforcement learning agents for searching feasible operating conditions of power grid that satisfy security constraints at base case and under contingencies. The detailed design corresponds to the multiple stages of manual tuning of operating conditions by power grid engineers when creating future operating scenarios. Through our testing conducted on actual power grid models, the proposed methodology exhibits initial success and effectiveness in relieving burdens of engineers, by automating such processes. We hope this work can be of help to the power community and promote applications of AI in power engineering.

## Acknowledgments and Disclosure of Funding

This work is funded by the SGCC Zhejiang Electric Power Company Science and Technology project, "Research on Power Grid Operation Scenarios Based on Deep Reinforcement Learning", under contract: SGZJJH00DKJS2000297.